\pdfoutput=1

\documentclass[11pt]{article}

\usepackage{naacl2021}

\usepackage{times}
\usepackage{latexsym}

\usepackage[T1]{fontenc}

\usepackage[utf8]{inputenc}
\usepackage{multirow}
\usepackage{subfig}
\usepackage{enumerate}
\usepackage{caption}
\usepackage{graphicx}
\usepackage{xspace}
\usepackage{booktabs}       
\usepackage{nicefrac}
\usepackage{paralist}
\usepackage{makecell}
\usepackage{todonotes}

\usepackage{microtype}

\usepackage{amsmath,amsfonts,bm}

\newcommand{\ie}{\textit{i.e.},\xspace}

\def\ie{{\em i.e.,}\xspace}
\def\cf{{\em c.f.,}\xspace}

\definecolor{mygray}{gray}{0.25}

\newcommand{\resource}[1]{\textsc{#1}}

\def\eqref#1{equation~\ref{#1}}

\newcommand{\tabref}[1]{Table~\ref{#1}\xspace}

\def\1{\bm{1}}

\def\vx{{\bm{x}}}
\def\vy{{\bm{y}}}

\DeclareMathAlphabet{\mathsfit}{\encodingdefault}{\sfdefault}{m}{sl}
\SetMathAlphabet{\mathsfit}{bold}{\encodingdefault}{\sfdefault}{bx}{n}

\title{
Model Extraction and Adversarial Transferability,\\Your BERT is Vulnerable!}

\author{Xuanli He$^\dagger$, Lingjuan Lyu$^\ddagger$, Qiongkai Xu$^\ast$, Lichao Sun$^\mathsection$\\
$^\dagger$ Monash University, xuanli.he1@monash.edu\\
$^\ddagger$ Ant Group, lyulingjuan.llj@antgroup.com
\\
$^\ast$ The Australian National University and Data61 CSIRO, Qiongkai.Xu@anu.edu.au \\
$^\mathsection$ Lehigh University, lis221@lehigh.edu
}

\begin{document}
\maketitle
\begin{abstract}
Natural language processing (NLP) tasks, ranging from text classification to text generation, have been revolutionised by the pretrained language models, such as BERT. 
This allows corporations to easily build powerful APIs by encapsulating fine-tuned BERT models for downstream tasks.
However, when a fine-tuned BERT model is deployed as a service, it may suffer from different attacks launched by the malicious users. 
In this work, we first present 
how an adversary can steal a BERT-based API service (the victim/target model) on multiple benchmark datasets with limited prior knowledge and queries.
We further show that the extracted model can lead to highly transferable adversarial attacks against the victim model.
Our studies indicate that the potential vulnerabilities of BERT-based API services still hold, even when there is an architectural mismatch between the victim model and the attack model. Finally, we 
investigate two defence strategies to protect the victim model, and find that unless the performance of the victim model is sacrificed, both model extraction and adversarial transferability can effectively compromise the target models.
\end{abstract}

\section{Introduction}
Recently, owing to the success of pretrained BERT-based models~\citep{devlin2018bert, liu2019roberta,sun2020mixup}, the downstream NLP tasks have been revolutionised in the form of the limited task-specific supervision via fine-tuning on BERT models.
Meanwhile, commercial task-oriented NLP models, built on top of BERT models, are often deployed as pay-per-query prediction APIs for the sake of the protection of data privacy, system integrity and intellectual property.

As publicly accessible services, commercial APIs have become victims of different explicit attacks, such as privacy attack~\citep{lyu2020differentially, lyu2020towards,7958568}, adversarial attack~\cite{shi2018generative}, etc. Recently, prior works have also found that with the aid of carefully-designed queries and outputs of the NLP APIs, many existing APIs can be locally imitated via model extraction~\citep{krishna2019thieves,wallace2020imitation}, which raises concerns of the vulnerability of NLP APIs. 
For instance, competing companies can 
imitate the victim model with a negligible cost. Since the considerable investment of data annotation and algorithm design are sidestepped, the competing companies would be able to launch an identical service with a more competitive price than the victim companies. Such security issue can be exacerbated, when the back-end pertained models, such as BERT, are publicly available~\citep{krishna2019thieves}.

Beyond model extraction, we further demonstrate the 
adversarial examples crafted by the extracted model could be transferred to the black-box victim model. 
From the perspective of commercial competition, if the competitors manage to predicate incorrect predictions of the victim services, they can launch an advertising campaign against the victim model with these adversarial examples.

In summary, we investigate the vulnerabilities of publicly available NLP classification APIs 
through a two-stage attack. First, a model extraction attack is issued to obtain a local copy of the target model. Then, we conduct adversarial attacks against the extracted model, which is empirically transferable to the target model. 
To patch these vulnerabilities, we mount two basic defence strategies on the victim models. The empirical results show that without corrupted predictions from the victims, model extraction and adversarial example transferability are resilient to the defence. 
Our results spotlight the risks of using pretrained BERT to deploy the APIs through the lens of model extraction attack and adversarial example transfer attack. 
Such attacks can be conducted at a cost of as little as \$7.1.\footnote{Code is available at \url{https://github.com/xlhex/extract\_and\_transfer}}

\section{Related Work}

\subsection{Model Extraction Attack (MEA)} 

Model extraction attacks (also referred to as ``model stealing") have been effectively applied to different tasks, ranging from computer vision tasks~\citep{orekondy2019knockoff} to NLP tasks~\citep{chandrasekaran2020exploring}.

In a nutshell, model extraction enables malicious users to forge the functionality of a black-box victim model as closely as possible. The activity seriously causes the intellectual property infringement.
Additionally, the follow-up attacks can be facilitated as the aftermath of the model extraction. Particularly, an adversarial attack can be built upon the extracted model, which is able to enhance the successful rate of fooling the victim model.

\subsection{Adversarial Transferability in NLP}

As a byproduct of the adversarial attack, it has been shown that adversarial transferability encourages a transition of the adversarial examples from one model to other models~\citep{liu2016delving,papernot2017practical}, especially in computer vision research. Although such property has been explored by a few recent works in NLP systems~\citep{sun2020adv,wallace2020imitation}, it remains largely unexplored for the BERT-based APIs, and whether the transferability could succeed when the substitute (extracted) model and the victim model have different architectures.

\begin{figure*}[t]
    \centering
    \includegraphics[width=0.99\textwidth]{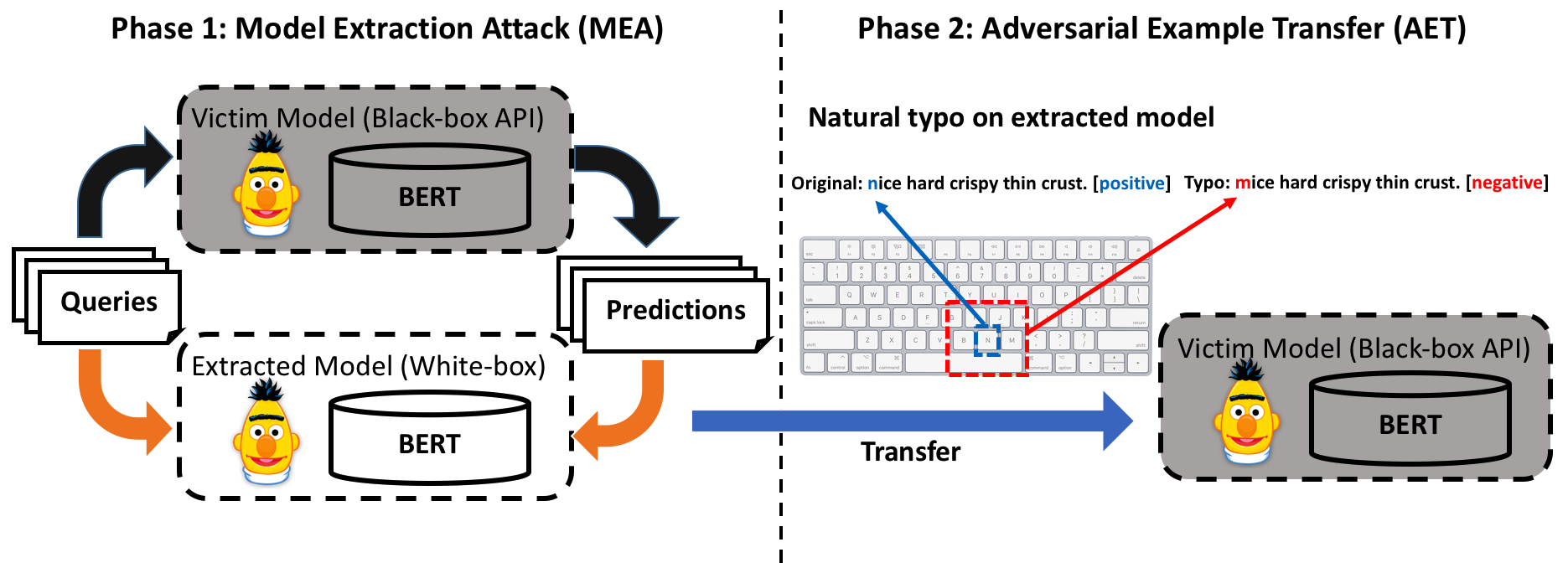}
    \vspace{-2mm}
    \caption{The workflow of the proposed attacks on BERT-based APIs. In phase 1, Model Extraction Attack (MEA) labels queries using the victim API, and then trains an extracted model on the resulting data. In phase 2, Adversarial Example Transfer (AET) generates adversarial typo examples on the extracted model, and transfers them to the victim API.}
    \label{fig:attack_pipeline}
    \vspace{-3mm}
\end{figure*}

\section{Attack on BERT-based API}

Our attacks against BERT-based APIs consist of two phases, \emph{Model Extraction Attack} (MEA) and \emph{Adversarial Example Transfer} (AET), as depicted in Figure~\ref{fig:attack_pipeline}.

\subsection{Model Extraction Attack (MEA)}
\label{sec:MEA}
In the first phase, we assume that a ``victim model'' $\mathrm{M_{v}}$ is commercially available 
as a prediction API for target task $\mathcal{T}$. 
An adversary attempts to reconstruct a local copy $\mathrm{M_{e}}$ (``extracted model'') of $\mathrm{M_{v}}$ via querying $\mathrm{M_{v}}$. 
Our goal is to extract a model with comparable accuracy to the victim model.
Generally, MEA can be formulated as a two-step approach, as illustrated by the left figure in Figure~\ref{fig:attack_pipeline}: 
\begin{enumerate}
\item 
Attackers craft a set of inputs as queries, then send them to the victim model (BERT-based API) to obtain predictions;

\item 
Attackers reconstruct a local copy of the victim model as an ``extracted model'' using the retrieved query-prediction pairs. 
\end{enumerate}

For each query $\vx_{i}$, $\mathrm{M_{v}}$ returns a $K$-dim posterior probability vector $\vy_{i} \in [0,1]^k$, with $\sum_k y_{i}^k=1$. The resulting dataset $\{\vx_{i}, \vy_{i}\}_{i=1}^{m}$ by $m$ queries is used to train $\mathrm{M_{e}}$. We assume that the attacker fine-tunes the public release of $f_{\text{bert}, \theta^*}$ on this dataset, with the objective of imitating the behaviour of $\mathrm{M_{v}}$.
Once the local copy of $\mathrm{M_{e}}$ is obtained, the attacker no longer needs to pay the original service provider.

\subsection{Adversarial Example Transfer (AET)}
 
In the second phase, we leverage the transferability of adversarial examples: we first generate adversarial examples for the extracted model, then transfer the generated adversarial examples to the victim model. The intuition of the experiment is based on the transferable vulnerabilities crossing the models -- the adversarial examples generated by the extracted model are transferable to the victim model. 
Here we use the extracted model to serve as a surrogate to craft adversarial examples in a white-box manner. Such attack aggravates the vulnerabilities of victim models.

\section{Experiments and Analysis}

\subsection{NLP Tasks and Datasets}

\begin{table}[t]
\small
    \centering
    \begin{tabular}{lllll}
      \toprule
    
Dataset &  \#Train & \#Dev& \#Test & Task  \\
  \midrule
{TP-US} &  22,142  & 2,767 & 2,767 & sentiment analysis\\
{Yelp} &  520K & 40,000 & 1,000 & sentiment analysis \\
{AG}  &  112K & 1,457 & 1,457 & topic classification\\
{Blog} &  7,098 &887 &887 & topic classification \\
     \bottomrule
 \end{tabular}
\vspace{-2mm}
    \caption{Statistic of sentiment analysis and topic classification datasets.}
    \vspace{-5mm}
    \label{tab:data}
\end{table}

To evaluate the efficacy of the proposed attacks, we select four NLP datasets covering two main tasks, i) sentiment analysis and ii) topic classification \cite{li2020text}. We use \resource{tp-us} from Trustpilot Sentiment dataset~\citep{hovy2015user} and \resource{Yelp} dataset \citep{zhang2015character} for sentiment analysis. We use \resource{ag} news corpus~\citep{del2005ranking} and Blog posts dataset from the blog authorship corpus~\citep{schler2006effects} for topic classification. We refer readers to Appendix \ref{sec:data_description} for more details about the pre-processing of these datasets.


\subsection{MEA Setup and Results}
\label{sec:mea_setup}

\paragraph{Attack Strategies:}

We assume that both victim and extracted models are initialised from a freely available pretrained BERT. Once the victim model is task-specifically fine-tuned by following Section~\ref{sec:MEA}, it can be queried as a black-box API. Afterwards, the extracted model can be obtained through imitating the victim model.
Following ~\citet{krishna2019thieves}, the queries start from the size of 1x to that of victim's training set, then scale up to 5x. We test the accuracy of the victim model and the extracted model on the same held-out set for a fair comparison.

\paragraph{Query Distribution:}

To examine the correlation between the query distribution ($D_A$) and the effectiveness of our attacks on the victim model trained on data from $D_V$ (\cf\tabref{tab:data}), we explore the following two different scenarios: 
(1) we use the same data as the original data of the victim model ($D_A=D_V$). Note that attackers have no true labels of the original data; (2) we sample queries from different distribution but same domain as the original data ($D_A \neq D_V$).

Since the owners of APIs tend to use the in-house datasets, it is difficult for the attacker to know the target data distribution as a prior knowledge. Therefore, our second assumption is closer to the practical scenario.
As the training datasets of the victims are sourced from either review domain or news domain, we consider datasets from these two domains as our queries. Specifically, we leverage Amazon review dataset~\citep{zhang2015character} or CNN/DailyMail dataset~\citep{hermann2015teaching} to query the victim models.

         

\begin{table}[t]
\small
    \centering
    \begin{tabular}{llccccc}
    \toprule
Model   &\#Q & TP-US & Yelp & AG  & Blog\\
         \midrule
Victim model & &  85.5 & 95.6 &  94.5 & 97.1\\
         \midrule
$D_A=D_V$ & &   \textbf{86.5} & \textbf{95.7}  &  \textbf{94.5}  & \textbf{96.8}\\
        \midrule
\multirow{2}{*}{{\makecell{$D_A \neq D_V$ \\(review)}} }
       & 1x  &   85.3 & 94.1 &  88.6 & 88.2 \\
        & 5x &  85.8 & 95.0 &  91.3 & 92.8 \\
        \midrule
\multirow{2}{*}{{\makecell{$D_A \neq D_V$ \\(news)}} }
         
        & 1x  &   84.2	& 91.1&  90.5 & 83.1 \\
        & 5x  &   85.5	& 93.1	& 92.3 & 87.6	 \\
        \bottomrule
    \end{tabular}
    \vspace{-2mm}
    \caption{Accuracy [\%] of the victim models and the extracted models among different datasets in terms of domains and sizes. \#Q: number of queries. 
    }
    \label{tab:mea}
    \vspace{-6mm}
\end{table}

According to \tabref{tab:mea}, we have observed that: 1) the success of the extraction correlates to the domain closeness between the victim's training data and the attacker's queries;
2) using same data even outperforms the victim models, which is also known as self-distillation~\cite{furlanello2018born}; 3) albeit the different distributions brought by review and news corpora, our MEA can still achieve 0.85-0.99$\times$ victim models' accuracies when the number of queries varies in \{1x,5x\}. Although more queries suggest a better extraction performance, small query budgets (0.1x and 0.5x) are often sufficiently successful. More results are available in Appendix \ref{app:ablation}. 
From now on, unless otherwise mentioned, we will use \textit{news} data for \resource{ag} news, and \textit{review} data for \resource{tp-us}, Blog and Yelp.\footnote{Empirically, we do not have access to the original training data of the victim model.}

\paragraph{Costs Estimation:}
We analyse the efficiency of MEA on various classification datasets. Each query is charged due to a pay-as-you-use policy adopted by service providers. We estimate costs for each task in \tabref{tab:cost} according to Google APIs\footnote{\url{https://cloud.google.com/natural-language/pricing}} and IBM APIs\footnote{\url{https://www.ibm.com/cloud/watson-natural-language-understanding/pricing}}. Considering the efficacy of model extraction, the cost is highly economical and worthwhile.

\begin{table}[t]
    \centering
    \begin{tabular}{llrr}
      \toprule
    
Dataset &  \#Query & Google price & IBM price  \\
  \midrule
{TP-US} & 22,142 & \$22.1 & \$66.3\\
{Yelp} & 520K  & \$520.0 & \$1,560.0\\ 
{AG}  &  112K & \$112.0 & \$336.0\\
{Blog} &  7,098 & \$7.1 & \$21.3\\
     \bottomrule
 \end{tabular}
    \caption{Estimate costs of model extraction on different datasets}
    \label{tab:cost}
\end{table}

\subsection{AET Setup and Results}
\label{sec:eat_setup}

After extracting 
a black-box victim model into a white-box extracted model, a white-box adversarial attack can be implemented. We first generate adversarial examples on the extracted model, then examine whether these examples are transferable to the target victim model. To evaluate such pseudo white-box attack, we assess it via a transferability metric, which refers to the misclassification rate of adversarial samples on the victim APIs.


To generate natural adversarial examples, we follow the protocol~\cite{sun2020adv} that leverages the gradients of the gold labels w.r.t the embeddings of the input tokens to find the most informative tokens, 
which have the largest gradients among all positions within a sentence. Then we corrupt the selected tokens with one of the following typos: 1) Insertion; 2) Deletion; 3) Swap; 4) Mistype: Mistyping a word though keyboard, such as “oh” $\rightarrow$ “0h”; 5) Pronounce: Wrongly typing due to the close pronounce of the word, such as “egg” $\rightarrow$ “agg”; 6) Replace-W: Replace the word by the frequent human behavioural keyboard typo based on the Wikipedia statistics~\cite{sun2020natural}.

\begin{table}[h]
\small
\begin{center}
\begin{tabular}{cccccc}
\toprule
&& TP-US& Yelp & AG & Blog\\

\cmidrule{2-6}
        \multirow{9}{*}{\rotatebox{90}{black-box}} &deepwordbug & & & &\\
         &1x & 18.4 & 18.5 &  25.6  &52.9\\
         &5x & 18.2 & 25.7 & 35.3 &67.8\\
\cmidrule{2-6}
         &textbugger  & & &  &\\ 
         &1x & 21.3 & 16.3 & 16.1& 41.2\\
         &5x & 21.1 &  21.3 & 24.7& 62.7\\
\cmidrule{2-6}
         &textfooler & & & &  \\
         &1x &  27.5 & 17.3  &  18.5 & 34.7\\ 
         &5x & 27.1  & 21.9  & 24.9 &64.4\\ 
\midrule
\multirow{3}{*}{\makecell{w-box \\ (ours)}} &adv-bert & & &  &\\
         &1x &   \textbf{48.6} & 35.5&  47.5 & 64.9\\
         &5x & 47.3 &  \textbf{43.3}  & \textbf{53.6} & \textbf{76.5}\\
\bottomrule
         
    \end{tabular}
    \end{center}
    \vspace{-2mm}
    \caption{Transferability is the percentage of adversarial examples transferred from the extracted model to the victim model. deepwordbug~\citep{gao2018black}; textbugger~\citep{li2018textbugger}; textfooler~\citep{jin2019bert}; adv-bert~\citep{sun2020adv}. w-box: white-box.}
    \label{tab:adv}
\end{table}

In order to understand whether our extracted model manages to improve the transferability, we also launch a list of black-box adversarial attacks in the same manner. \tabref{tab:adv} demonstrates that our pseudo white-box attack makes the victim model more vulnerable to adversarial examples in terms of transferability — more than twice effective in the best case, compared to the black-box counterparts.
This corroborates our claim that the extracted model, retaining a high-fidelity imitation of the victim model, severely impairs the output integrity of the victim model, indicated as the considerable increase of the transferable examples.

In general, \tabref{tab:adv} also shows that more queries (5x {\em v.s.}1x) lead to better attack performances. We believe this conspicuous gain attributes to the higher fidelity to the victim model, obtained by a better extraction (\cf~\tabref{tab:mea}).


\begin{table}
\begin{center}
\begin{tabular}{ llll } 
 \toprule
 Victim & Extracted &  MEA & AET  \\
 \midrule
 BERT-large & BERT-large & \textbf{91.0} & \textbf{59.3}\\
 BERT-base & BERT-large &  90.7& 37.2\\
 BERT-base & BERT-base & 90.5 & 47.5\\
 BERT-large & BERT-base &    89.9&42.7 \\
 \bottomrule
\end{tabular}
\end{center}
\vspace{-3mm}
\caption{
Attack performance on AG news with mismatched BERT architectures. 
}
\label{tab:bert_mismatch}
\vspace{-5mm}
\end{table}



\begin{table*}[ht!]
\small
    \centering
    \begin{tabular}{lcccccccc}
\toprule
\multirow{2}{*}{} &\multicolumn{2}{c}{\resource{tp-us}} &\multicolumn{2}{c}{Yelp} &\multicolumn{2}{c}{\resource{AG}} & \multicolumn{2}{c}{Blog}\\
\cmidrule(lr){2-3}  \cmidrule(lr){4-5} \cmidrule(lr){6-7} \cmidrule(lr){8-9}
       & MEA $\downarrow$  &AET $\downarrow$    & MEA $\downarrow$ &AET $\downarrow$  & MEA $\downarrow$  &AET $\downarrow$ & MEA $\downarrow$  &AET $\downarrow$  \\
\midrule
\resource{No def.} & 85.3 (85.5) &  48.6  & 94.1 (95.6) & 35.5& 90.5 (94.5) & 47.5 & 88.2 (97.1) & 64.9\\
\midrule
\resource{soft.} ($\tau$=0.0) & 84.6 (85.5) &  40.2  & 93.7 (95.6) & 21.6 &  90.0 (94.5) &   33.0 & 85.6 (97.1) & 51.4\\
\resource{soft.} ($\tau$=0.5) & 85.1 (85.5) & 50.9  & 93.8 (95.6) & 20.6 & 90.3 (94.5) &  33.1 & 85.7 (97.1) & 61.5\\
\resource{soft.} ($\tau$=5.0) & 85.3 (85.5) & 58.7  & 94.5 (95.6)& 36.1 &  90.9 (94.5)&  53.3 & 86.7 (97.1) & 66.7\\
\midrule
\resource{pert.} ($\sigma$=0.05) & 85.3 (85.5)	& 55.0 & 93.9 (95.6)& 29.2 &90.1 (94.3)	& 40.3 & 85.9 (96.2) & 64.0\\
\resource{pert.} ($\sigma$=0.20) & 85.1 (85.4)	& 49.7 & 93.7 (95.5) & 25.4	& 90.2 (94.3)	& 35.4 & 85.3 (95.4) & 52.2\\
\resource{pert.} ($\sigma$=0.50) & \textbf{82.7} (63.2)	&  \textbf{28.3}	& \textbf{92.5} (87.8) & \textbf{16.6} & \textbf{89.0} (76.4)	& \textbf{20.0} & \textbf{81.8} (62.2) & \textbf{32.8}\\
\bottomrule

\end{tabular}%

\vspace{-2mm}
\caption{Attack performance under different defences (\resource{no def.}, \resource{soft.} and \resource{pert.}) and datasets. 
\textbf{Lower} scores indicate better defences. 
All experiments are conducted with 1x queries. Numbers in parentheses are accuracy of victim models with defence.}
\vspace{-5mm}
    \label{tab:defense}
\end{table*}

\subsection{Architecture Mismatch}
\label{sec:arch_diff}
In practice, the adversary may not know the victim's model architecture. {\color{black}Hence we also study the attacking behaviours under the different architectural settings.} 
According to \tabref{tab:bert_mismatch}, when both the victim and the extracted models adopt BERT-large, the vulnerability of the victim is magnified in all attacks, which implies that the model with higher capability is more vulnerable to our attacks. As expected, 
the efficacy of AET can be alleviated when an architectural mismatch exists.\footnote{More experiments can be found in Appendix \ref{app:arch}}

\section{Defence}
\label{sec:defense}
We next briefly discuss 
two defence strategies the victim model can adopt to counter these attacks.

\begin{itemize}
    \item \textbf{Softening predictions (SOFT).} A temperature coefficient $\tau$ on softmax layer manipulates the posterior probability distribution. A higher $\tau$ leads to smoother probability, whereas a lower one produces a sharper distribution. When $\tau$=0, the posterior probability becomes a hard label.
    \item \textbf{Prediction perturbation (PERT).} Another defence method is adding normal noise with variance $\sigma$ to the predicted probability distribution. The larger the variance of the noise distribution, the stronger the defence.
\end{itemize}

\tabref{tab:defense} indicates that varying temperature on softmax cannot defend the victim model against MEA, except for $\tau$=0 (hard label), which can degrade all attacks to some extent.

Regarding perturbation, it can achieve a significant defence at the cost of the accuracy of the victim models. Surprisingly, when $\sigma$=0.50, MEA surpasses the victim model. 
We conjecture that albeit the perturbed post-softmax probability, the extracted model can still acquire certain informative knowledge via model extraction. We will conduct an in-depth study on this in the future.

To sum up, both MEA and AET pose severe threats to the BERT-based APIs, even when the adversary merely has access to limited or erroneous predictions.

\section{Conclusions}
This work goes beyond model extraction from BERT-based APIs, and we also identify the extracted model can largely enhance adversarial example transferability even in difficult scenarios, \ie limited query budget,  queries from different distributions, or architectural mismatch. 
Extensive experiments based on representative NLP datasets and tasks under various settings demonstrate the effectiveness of our attacks against BERT-based APIs. 
In the future, we plan to extend our work to more complex NLP tasks, and develop more effective defences.

\section*{Acknowledgements}
We would like to thank anonymous reviewers for their valuable feedback and constructive suggestions. The computational resources of this work are supported by the Multi-modal Australian ScienceS Imaging and Visualisation Environment (MASSIVE) (\url{www.massive.org.au}).

\bibliography{anthology,custom}
\bibliographystyle{acl_natbib}

\newpage
\appendix

\section{Dataset Description}
\label{sec:data_description}
\paragraph{Trustpilot~(TP).} Trustpilot Sentiment dataset \citep{hovy2015user} contains reviews associated with a sentiment score on a five point scale.
The original dataset is comprised of reviews from different locations, however in this paper, we only derive \resource{tp-us} for study.

\paragraph{AG news.} We use \resource{ag} news corpus~\citep{del2005ranking}. This task is to predict the topic label of the document, with four different topics in total. Following \citep{zhang2015character, jin2019bert}, we use both ``title'' and ``description'' fields as the input document.

\paragraph{Blog posts (Blog).} We derive a blog posts dataset (Blog) from the blog authorship corpus presented~\citep{schler2006effects}.
We recycle the corpus preprocessed by ~\citet{coavoux2018privacy}, which covers 10 different topics.

\paragraph{Yelp Polarity (Yelp).} Yelp dataset is a document-level sentiment classification \citep{zhang2015character}. The original dataset is in a five point scale (1-5), while the polarised version assigns negative labels to the rating of 1 and 2 and assigns positive labels to 4 and 5.  

\section{Training Details}
We use Huggingface~\cite{wolf-etal-2020-transformers} as the codebase. Each model is trained for 4 epochs on a NVIDIA V100 GPU, with a batch size of 64. We use AdamW~\citep{loshchilov2018decoupled} with a learning rate of 5e-5.

\section{Performance of Different Query Size}
\label{app:ablation}
Due to the budget limit, the attacker cannot issue massive requests. To investigate the attack performance of model extraction under the low-resource setting, we conduct two additional experiments, which only utilise 0.1x and 0.5x of the training data of the victim models respectively. According to \tabref{tab:mea_full}, the overall performance of extracted models is comparable to the victim models. Only Blog with 0.1x training suffers from a drastic drop, as Blog uses the least number of training samples in all four datasets. In addition, distant domains exhibit significant degradation, when compared to the close ones. For example, sampling 0.1x-5x queries from news data present a more stable attack performance against the victim model trained with AG news than Blog. 

\begin{table}[t]
\small
    \centering
    \begin{tabular}{cc|ccccc}
    \toprule
       & \#Q &  AG  & Blog & TP-US & Yelp\\
         \midrule
     victim model &  &94.47 & 97.07 &  85.53 & 95.57\\
         \midrule
  $D_A=D_V$ & &  94.54  & 96.77 &  86.48 & 95.72 \\
        \midrule
   
   \multirow{4}{*}{{\makecell{$D_A \neq D_V$ \\(review)}} } 
         &0.1x &  86.57 &  36.83& 79.95  & 92.39 \\
        &0.5x &  87.31 &84.59  & 84.21  & 93.25\\
       &1x  &  88.63 & 88.16 &  85.33 & 94.06  \\
        &5x &   91.27 & 92.75 &  85.82 & 94.95 \\
        \midrule
    \multirow{4}{*}{{\makecell{$D_A \neq D_V$ \\(news)}} } 
          &0.1x &  89.13 &18.04  & 79.20  & 88.24\\
        &0.5x &  89.84 & 32.92 & 84.18  & 89.76\\
        &1x  &   90.48 & 83.13	&  84.15	& 91.06 \\
        &5x  &   92.26 & 87.64	& 85.46	& 93.13\\
        \bottomrule
    \end{tabular}
    \caption{Accuracy [\%] of the victim models and the extracted models among different datasets in terms of domains and sizes. \#Q: number of queries. 
    }
    \label{tab:mea_full}
\end{table}

\section{Architectural Mismatch}
\label{app:arch}
In~\tabref{tab:arch_mismatch}, we experiment with different models, including BERT~\citep{devlin2018bert}, RoBERTa~\citep{liu2019roberta} and XLNET~\citep{yang2019xlnet}. Although the architectural difference can cause some drops in MEA and AET, overall the proposed attacks are still effective.


\begin{table}
\begin{center}
\begin{tabular}{ llll } 
 \toprule
 Victim & Extracted &  MEA & AET  \\
 \midrule
 BERT-large & BERT-base &    89.88&42.7 \\
 RoBERTa-large & BERT-base & 89.74 & 27.7\\
 RoBERTa-base & BERT-base & 89.45 & 36.4\\
 XLNET-large & BERT-base &89.66 & 32.7\\
 XLNET-base & BERT-base & 89.27 & 34.4\\
 \midrule
 BERT-base & BERT-base & 90.48 & 47.5\\
 \bottomrule
\end{tabular}
\end{center}
\caption{
Attack performance on AG news with mismatched BERT architectures. 
}
\label{tab:arch_mismatch}
\end{table}

\end{document}